# Spell Correction for Azerbaijani Language using Deep Neural Networks

**Ahmad Ahmadzade**
Kapital Bank
Risk Management Department
Baku, Azerbaijan
Ahmad.Ahmadzada@kapitalbank.az

**Saber Malekzadeh**
Khazar University
Lumos Extension School of Data
Baku, Azerbaijan
saber.malekzadeh@sru.ac.ir

*Abstract*— Spell correction is used to detect and correct orthographic mistakes in texts. Most of the time, traditional dictionary lookup with string similarity methods is suitable for the languages that have less complex structure such as English language. However, Azerbaijani language has more complex structure and due to morphological structure of it, derivation of words is plenty that several words are derived from adding suffices, affixes to the words. Therefore, in this paper sequence to sequence model with attention mechanism is used to develop spelling correction for Azerbaijani. Total 12000 wrong and correct sentence pairs used for training and model is test on 1000 real world misspelled words and F1-score results are 75% for distance 0, 90% for distance 1, and 96% for distance 2.

## I. Introduction

We all know, how data is important today. However, correctness and clarity of the transmitted information is more important. Most of the orthographical errors are found in transmitted information on social media, chat applications, and emails. Spelling correction is also important in the search engines and translation systems. Expressions and meaning of sentences that are made up the context is very important. It is not pleasant to read a content that has many grammatical and spelling mistakes.

The main reason of the research is to develop a spelling checker algorithm is that users who try to use Azerbaijani in their blogs, social media have no way to be sure about spelling for the texts they have written. That is why, there is a necessity to create new algorithm to check orthographical mistakes. It is observed that majority of Azerbaijani writers do not use some of our letters "ö ğ ı ə ç ş" because they are writing the text in English keyboard. So main part of this project will not only be catch misspelling words it also corrects the letters the users add with English keyboard.

In order to develop effective spelling correction, typing behavior of the people is studied to find what kind of mistakes and errors they make when they write. For this purpose, Azerbaijani comments on Facebook is examined. Azerbaijani is an agglutinative language so different languages that have complex morphological structure are studied. One of the problems of this research is scarcity of the labeled Azerbaijani sentences. Incorrect sentences and their correct version (labels) are needed in order to apply advance techniques such as deep neural networks. In order to solve data insufficiency, different methods are used such as changing correct sentences into incorrect sentences by inserting, deleting, transposing letters to sentences and other data augmentation techniques are applied.

There are a lot of researchers that have been dealing with spelling checking and introduced tradition techniques such as dictionary lookup and distance metrics for spelling correction.

Levenshtein distance algorithm is used which is one of the string similarity checking algorithm that uses edit distance to measure the similarity between two strings [1] [2]. Easiest way for spell checking is to construct a word dictionary or corpus that contains all the correctly written words and use string similarity checking algorithm to find most similar words from dictionary to the inputted word [3] [4]. Usual dictionary lookup methods take long time, because it matches the given words with every word from the dictionary. If you have a sentence and it has five words and your dictionary word corpus have 500 thousand words, the complexity of similarly checking algorithm will be number of inputted words multiple number of words in the dictionary word corpus [5].

More advanced method for dictionary method is discussed. Beside from usual dictionary lookup, there is a method which is Lexicon lookup. Lexicon is a kind of data structure that stores words based on their alphabetic similarity in the form of tree structure. This algorithm is optimized for fast searching in the Tree [6] [7]. Another solution was provided which is improved version of Levenshtein distance algorithm is Damerau- Levenshtein distance. Beside from insertion, deletion, and substitution operations, Damerau added one additional operation which is transposition [8] [9]. Transposition costs for two successive letter pairs are required to get better results while using Damerau -distance algorithm [10] [11].

There are different probabilistic models that used for word corpuses or sequences. Models the assign the probabilities to the words are called as Language model [12] [13]. Instead string similarity, this model uses occurrence of the inputted word in the given context. One of the language models is N-gram. N-gram consists of words that can be mono-gram, bi-gram, (two words), tri-gram, (three words) and etc. N-gram model finds the probability of last word that is given in n-gram pairs.

In the following parts, related works are presented which introduce different solutions with neural networks and sequence to sequence modelling for similar problems as spelling correction. After related work, research methods are discussed which are about data collection, data pre-processing, structure of sequence to sequence model, loss function and attention mechanism. After that proposed model and results are discussed.

## II. RELATED WORK

Many researchers have presented several solutions for sequence to sequence problems where input and output of the neural network model is sequence and difference hybrid methods are presented for spelling correction.

For data scarcity, experiments are performed on synthetic datasets created for Indic languages, Hindi and Telugu, by incorporating the spelling mistakes committed at character level and sequence-to-sequence deep learning model is used [14]. Spelling corrects consists of two steps. The error detection and error correction. The detection employs a neural network with LSTM layers which is used to find incorrect words and their positions. Error correction steps generate candidate words for incorrect words based on probability [15].

The use of a fixed-length vector is a pivot point in improving the performance of this basic encoder–decoder architecture and they propose a method extend this architecture by allowing a model to automatically search for parts of a source sentence that are relevant to predicting a target word, without having to form these parts as a hard segment explicitly. They achieved high performance for the English-to-French translation [16].

Bi-Directional Attention Flow network is applied, and attention flow mechanism is used to get context aware representation of query without early summarization. Their experiment achieved good results on Stanford Question Answering Dataset (SQuAD) and CNN/DailyMail cloze test [17].

A better method is proposed which its approach significantly improves on state-of-the-art statistical language models which are probabilistic models and that the proposed approach allows to take advantage of longer contexts [18]. Local and global attention mechanism is used for machine translation. Their ensemble method that used different attention architectures got high improvement for English-German translation task [19].

Autoencoder neural network architecture with bidirectional LSTM with attention mechanism is used for Azerbaijani spelling correction. Instead of simulated data, manually labeled words are used for training. Although proposed model speed up traditional dictionary lookup process and corrects a word at a time, it can be improved so that it can correct sentences at a time considering contextual meanings [20].

## III. PROPOSED MODEL

In this section, data collection, pre-processing, tokenization and padding techniques, final proposed model, and results are presented.

### A. Data Collection and Pre-Processing

For spelling correction, wrong and correct sentence pairs are needed to feed into deep neural network. Due to scarcity of labeled data, data augmentation technique is used to generate synthetic data from original data. Wrong sentences are generated from correct sentences by applying different operations such as insertion, deletion, transposition, and substitution.

These operations are randomly performed on each word in the sentence. Insertion, deletion, and transposition operations are performed less than 40 percent of time. However, for the substitution operation, more intelligent way is used. As disused in introduction, people use some English letters (e, w, c, g) instead of Azerbaijani letters (ə, ş, ç, ğ). So, instead of substituting a letter with random letter, it can be replaced with a letter that is confused and mistyped most of them time. In order to analyze mistyping behavior of people, comments of people on different pages in Facebook are scraped and analyzed. In table 1, letters and their most misspelled and confused letters are given.

Table 1. Correct letters and their most mistyped versions.

| Correct letter | Misspelled letters |
|---|---|
| 'ə' | ['e', 'i', 'a', 'ı', 'ə'] |
| 'ş' | ['w', 's', 'sh', 'i'] |
| 'ç' | ['c', 'ch'] |
| 'q' | ['g', 'x', 'k', 'ğ'] |
| 'ı' | ['i', 'l', 'u', 'ə', 'o', 's', 'a'] |
| 'r' | ['t', 'l', 'z', 'x'] |
| 'ğ' | ['g', 'q'] |
| 'l' | ['d', 'n', 'o'] |
| 'd' | ['n', 't', 'x'] |
| 'u' | ['h', 'i', 'ı', 'o'], |
| 'ö' | ['o', 'i', 'ü', 'e'] |
| 'ü' | ['u', 'i', 'e'] |
| 't' | ['d', 'r'] |
| 'y' | ['g', 't', 'j'] |
| 's' | ['z', 'c', 'd', 'ss'] |
| 'z' | ['x', 's'] |
| 'g' | ['q', 'y', 'k', 'ğ'] |
| 'n' | ['m', 'v', 'l'] |
| 'i' | ['u', 'y', 'ı'] |
| 'a' | ['i', 'u', 's', 'e', 'ə', 'm', 'ı', 'v'] |
| 'k' | ['y', 'g', 'q'] |

As substitution operation is done with logical ways than random letters, it is more preferred and used 60 percent of time. By generating synthetic words with this way, it becomes very close to real word errors, misspelling that people do. In table 2, two sample wrong, correct sentences pair provided. Here, each synthetic word is generated from the correct words in the sentence. Total of 12000 sentences pairs are creates in such way.

Table 2. Correct sentence and generated synthetic sentence.

| Sentence type | Sentence |
|---|---|
| Correct sentence | "Koronavirusla ən effektiv mübarizə aparan on ölkə arasında üçüncü yeri tutub." |
| Generated synthetic sentence | "Koronavirusla ən effeytiv mubaruzə aparan on onye arasında üchüncu eri tutqub." |
| Correct sentence | "Turizm sahəsində daha ciddi uğurlardan və nəticələrdən danışmaq olardı" |
| Generated synthetic sentence | "Turiz sahesinde dah cuddi ugurlardan və nəticələrdDən danışmaHq olardı" |

## B. Tokenization and Padding

Wrong and correct sentence pairs are in string format and to feed data to the model, sentences should be tokenized into either words or characters. TensorFlow provides great functions to work with text data. One is "Tokenizer" which convert text into word or character tokens, and it has different input parameters such as filters which removes specified characters from the text.

Azerbaijan language has rich morphological structure and a new word can be derived from a word by adding suffix, prefix, or infix. So, tokenizing a sentence into word token is bad idea, because tokenizer can only vectorize specified unique words in text corpus, not all words. For this problem, instead of tokenization a sentence into words, the sentence is tokenized into character token which can represent all possible words in a sentence. For each sentence, "<" sign is added to beginning of the sentence which defines start point of a sequence and ">" sign is added to the end of the sentence which defines stopping point of the sequence.

Proposed model requires same length sequence to be fed. So, max length of sentence is set to 20 which is average sentence length and padded with "post" which pads with value zero end of the sequence. Truncating is set to "pre" which removes elements from the beginning of the sequence if greater than max length (20). Two tokenizers are created, one for input sequence (wrong sentences) and another for output sequence (correct sentences).

All tokenized sentences should be in same sequence length to feed into the model. So, padding is applied to each sentence. Padding allows to set max length for the input sequence and provides "pre/post" padding.

## C. Autoencoder

In sequence-to-sequence tasks, most of time, encoder-decoder model is implemented by using Recurrent Neural Networks (RNN). RNNs are a type of Neural Network (NN) that in current step, value from previous step is used. In spelling correction problem, to predict next character, previous characters play crucial role. So, RNN has a feature which is Hidden state. It has a memory which keeps information from previous steps.

Encoder network can consist of several layers such as Embedding and recurrent layers as Gated Recurrent Unit (GRU) and Long-Short Term Memory (LSTM). The role of Encoder part is to converting input sentence (sequence of tokens) to feature vectors. These hidden features represent information from each element in input sequence.

Decoder part uses output of Encoder as an initial hidden state, current word or character and its own hidden state, in order to predict next word or character. In figure 1, General architecture of Encoder-Decoder model is illustrated.

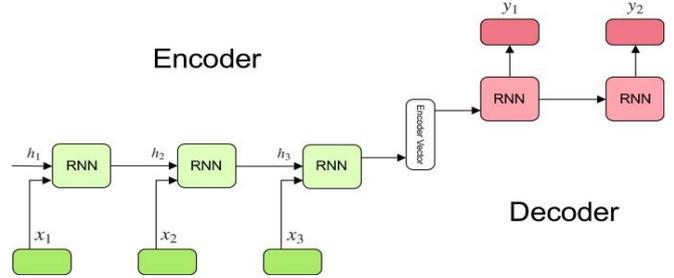

Figure 1. Encoder-Decoder architecture.

## D. Attention Mechanism

One of possible problem of simple Encoder-Decoder model it that final state of Encoder model may not produce all valuable information to the Decoder model. Especially, if the input is a long sentence, RNN may not pass all information from previous steps till the end of sentence. In order to solve this problem, attention mechanism is used between Encoder and Decoder to pass information effectively.

The goal of attention mechanism is to generate context vectors, where each vector carries information about how attention should be paid for each element in the sequence.
In figure 2, general structure of attention mechanism is shown which consists of 3 parts – Feed forward neural network, softmax calculation, and generation of context vector.

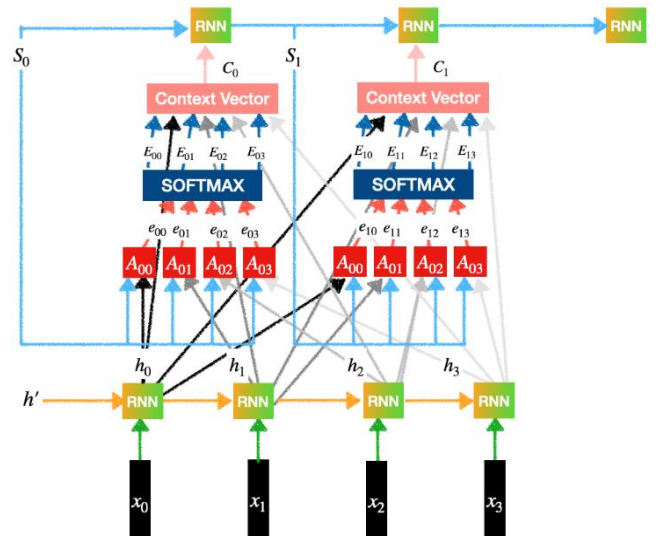

Figure 2. Attention mechanism

Feed forward neural networks ($A_{00}, A_{01}, A_{02}, A_{03}$) get input from previous decoder state and output of encoder state to generate outputs($e_{00}, e_{01}, e_{02}, e_{03}$). Each output is

calculated as in (1) where g is an activation function (sigmoid, ReLu etc.)

$$e_{0i} = g(S_0, h_i) \quad (1)$$

Softmax calculation as in (2) is performed to get attention weights that carries information for each element in the sequence.

$$E_{0i} = \frac{\exp(e_{0i})}{\sum_{i=0}^{3} \exp(e_{0i})} \quad (2)$$

After softmax calculation, context vector is generated as in (3) which is the product of probability distribution and output of encoder.

$$C_i = E_{i0*}h_0 + E_{i1*}h_1 + E_{i2*}h_2 + E_{i3*}h_3 \quad (3)$$

Context vectors are used in Decoder layer and it produces output sequence.

*E. The proposed model for Spelling Correction*

The proposed model is character-based sequence to sequence model where each wrong, correct sentence pair are tokenized into character token and fed into model. Model consists of 3 parts – Encoder, Decoder, and Attention mechanism. Previously, functions of each part are described in detail. In proposed model, the both encoder and decoder model consists of embedding and LSTM layers. Attention mechanism generates context vector which carries information from Encode to Decoder.

Embedding dimension for embedding layer is 500 and units for LSTM layer is 500. Data is fed into model as batches with size of 40. For the attention, bathdanau attention mechanism is used. It acts as additive attention with uses linear combination of encoder and decoder state. For the optimizer, Adam and for the loss function, Sparse Categorical Cross entropy is used. Architecture of the proposed model is showed in Figure 3 and detailed information about its layers and parameters in given in Figure 4.

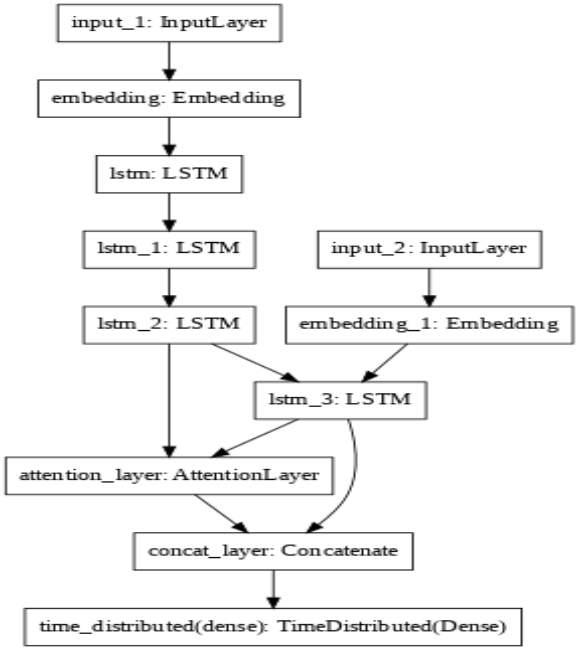

Figure 3. Visual architecture of the proposed model

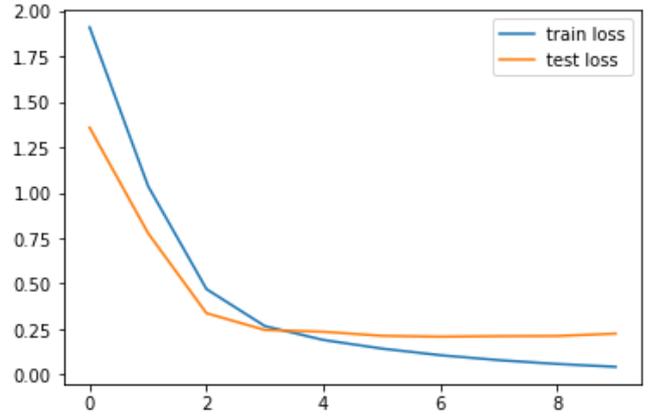

Figure 4. Detailed architecture of the proposed model

Model is trained in GPU enabled Google Collaboratory Python notebook with 10 epochs. Train and test loss of model is shown in Figure 2.

Figure 5. Train loss by epochs

Proposed model is validated on total of 3000 real world misspelled words and accuracy is measured by checking Levenshtein edit distance [1] between predicted words and real words. Distance 0 means model predicted the word very correctly. Distance greater 0 means that model predicted the words somehow correct but with some edit distance. As shown in figure 3, model accuracy increases if higher edit distance is chosen. Proposed model is also compared with the results of another author which is shown in table 3 [20].

Table 3. Accuracy of proposed and compared model

| Distance\Model | Proposed model | Compared model [20] |
|---|---|---|
| 0 | 75% | 53% |
| 1 | 90% | 80% |
| 2 | 96% | 92% |
| 3 | 98% | 98% |

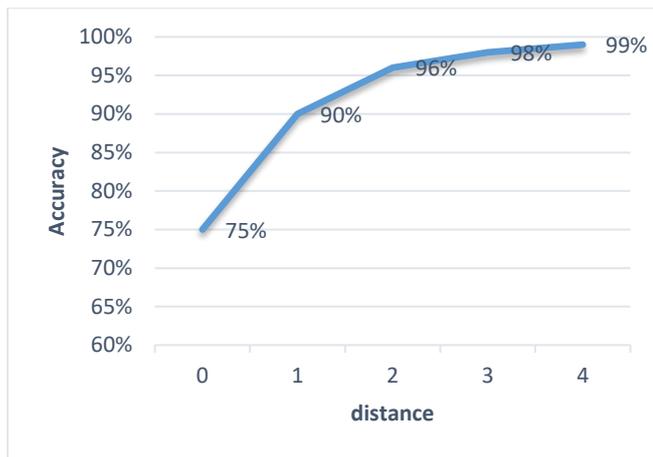

Figure 6. Model accuracy by edit distance.

## IV. CONCLUSION

In this research, encoder-decoder model with attention mechanism is applied to spelling correction for Azerbaijani language. Train and test data consist of sequences of incorrect and correct sentence pairs where incorrect sentences in train data is generated from correct sentences. Model is test on real word data and the overall results are 75% for edit 0, 90% for edit 1, and 96% for edit 2.